\def\year{2018}\relax
\documentclass[letterpaper]{article} 
\usepackage{aaai18}  
\usepackage{amsmath}
\usepackage{amsfonts}
\usepackage{amssymb}
\usepackage{mathtools}
\usepackage{todonotes}
\usepackage[noend]{algpseudocode}
\usepackage{algorithm}
\usepackage{bbm} 
\usepackage{graphicx}
\usepackage{hyperref} 
\usepackage{mwe}
\usepackage{caption}
\usepackage{tikz}
\usetikzlibrary{calc,tikzmark}

\algnewcommand\algorithmicinput{\textbf{Input:}}
\algnewcommand\INPUT{\item[\algorithmicinput]}
\algnewcommand\algorithmicoutput{\textbf{Output:}}
\algnewcommand\OUTPUT{\item[\algorithmicoutput]}
\usepackage{times}  
\usepackage{helvet}  
\usepackage{courier}  
\usepackage{acronym}
\usepackage{url}  
\usepackage{graphicx}  
\usepackage{hyperref}

\newcommand{\dist}{\mathbf{D}}

\newcommand{\traj}{\mathcal{T}}

\newcommand{\expect}{\mathbb{E}}
\newcommand{\ie}{{\it i.e.}}
\newcommand{\reals}{\mathbb{R}}
\newcommand*{\eg}{e.g.\@\xspace}

\acrodef{mdp}[MDP]{Markov decision process}
\acrodef{irl}[IRL]{Inverse Reinforcement Learning}
\acrodef{em}[EM]{Expectation-Maximization}
\acrodef{dpm}[DPM]{Dirichlet process mixture}
\acrodef{crp}[CRP]{Chinese Restaurant Process}
\acrodef{rl}[RL]{Reinforcement Learning}
\newtheorem{problem}{Problem} 

\begin{document}
%

\title{Inverse Reinforce Learning with Nonparametric Behavior
  Clustering}

\author{ Siddharthan Rajasekaran, Jinwei Zhang, and Jie Fu\\
Department of Electrical and Computer Engineering\\
Robotics Engineering Program\\
Worcester Polytechnic Institute\\
 sperundurairajas, jzhang8, jfu2 @wpi.edu
}
\maketitle

\begin{abstract} Inverse Reinforcement Learning (IRL) is the task of learning a single reward function given a Markov Decision Process (MDP) without defining the reward function, and a set of demonstrations generated by humans/experts. However, in practice, it may be unreasonable to assume that human behaviors can be explained by one reward function since they may be inherently inconsistent. Also, demonstrations may be collected from various users and aggregated to infer and predict users' behaviors. In this paper, we introduce the Non-parametric Behavior Clustering IRL algorithm to simultaneously cluster demonstrations and learn multiple reward functions from demonstrations that may be generated from more than one behaviors. Our method is iterative: It alternates between clustering demonstrations into different behavior clusters and inverse learning the reward functions until convergence. It is built upon the Expectation-Maximization formulation and non-parametric clustering in the IRL setting. Further, to improve the computation efficiency, we remove the need of completely solving multiple IRL problems for multiple clusters during the iteration steps and introduce a resampling technique to avoid generating too many unlikely clusters. We demonstrate the convergence and efficiency of the proposed method through learning multiple driver behaviors from demonstrations generated from a grid-world environment and continuous trajectories collected from autonomous robot cars using the Gazebo robot simulator.   
\end{abstract}

\section{Introduction}

\ac{irl} provides a structured way of learning a reward function to obtain the required complex behavior using demonstrations from experts and the model of the system \cite{abbeel2004apprenticeship,ng2000algorithms,ziebart2010modeling}. It aims to answer ``What possible reward function could have made the expert choose these actions?". This is analogous to figuring out the intent behind the motions of the expert. Recently, several \ac{irl} algorithms have been developed \cite{abbeel2004apprenticeship,russell1995modern,ziebart2008maximum,ramachandran2007bayesian}. Among these algorithms, Maximum Entropy (MaxEnt) \ac{irl} \cite{ziebart2008maximum} employs the principle of maximum entropy to resolve noise and sub-optimality in demonstrated behaviors, which is the case with practical human behavior data. Recently, \ac{irl} has been applied to many practical applications including user preference of public transit from data \cite{wu2017datadriven}, inferring taxi driving route preferences \cite{ziebart2008maximum}, and demonstrating complex tasks to robots \cite{finn2016guided}.

However, variance in demonstrations may not be simply explained as
noise or sub-optimality.  On the one hand, the behavior data may be collected from multiple users and aggregated for inference. For
example, data aggregation in urban planning is needed due to data
scarcity, \eg, routes selected by a traveler over a week may not
constitute sufficient statistics. On the other hand, human behaviors
may not be explained by one reward function since they may be
inherently inconsistent. It is often impractical for a human to
consistently demonstrate a task due to, \eg, physiological
limitations, and short attention span. In this paper, we consider that
variance in demonstrations may be due to more than one cluster of
behavior, and different clusters correspond to different reward functions.
These clusters may also hint at unmodeled reward feature which, if
modeled, would better explain all the demonstrations. To this end, we
aim to develop an algorithm that automatically infers the clusters and
their corresponding reward functions.


The problem of inverse learning multiple reward functions from demonstrations has been studied in recent works \cite{nguyen2015inverse,babes2011apprenticeship,michini2012bayesian}. In general, two classes of algorithms have been developed: Parametric \cite{babes2011apprenticeship} and Nonparametric ones \cite{michini2012bayesian}. In \cite{babes2011apprenticeship}, the authors derived an \ac{em} approach that can cluster trajectories by inferring the underlying intent using \ac{irl} methods. The method assumes the number of clusters in the demonstrations are known and alternates between the E-step to estimate the probability of a demonstration coming from a cluster, for all combinations of trajectories and clusters, and the M-step to solve several \ac{irl} problems for multiple clusters. Without the knowledge of the number of clusters, \cite{michini2012bayesian} proposed to integrate nonparametric clustering with Bayesian \ac{irl} \cite{ramachandran2007bayesian} with multiple reward functions. The inference uses the Metropolis-Hasting (MH) algorithm that updates the cluster assignment and the reward function distributions in alternation. This method has been extended in \cite{shimosaka2015predicting} to infer driving behaviors given diverse driving environments, characterized by road width, number of lanes, one-or two-way street.

This paper develops an algorithm based on \ac{em} and nonparametric clustering for inverse learning multiple reward functions with automatic clustering of demonstrations. Similar to the \ac{em}-based \ac{irl}\cite{babes2011apprenticeship}, we propose to perform soft clustering on the demonstrated dataset. That is, instead of associating a demonstration with a single behavior, we have a probability that a particular demonstration comes from a particular behavior cluster. We use the EM algorithm to associate each demonstration to a distribution of clusters while simultaneously learning their reward functions in the inner loop. At the outer loop, we employ nonparametric clustering to associate demonstrations with clusters and generate new cluster with a nonzero probability to accommodate variance in the demonstrations.

Compared to the existing methods \cite{babes2011apprenticeship,michini2012bayesian}, our approach has several advantages. 
First, our method does not solve multiple \ac{irl} problems completely during each iteration in the iterative clustering and \ac{irl} procedure, unlike \cite{babes2011apprenticeship}. This allows us to achieve computational efficiency since
\ac{irl} involves solving \ac{rl} problems multiple times and thus is computationally intensive. 
Second, a resampling technique is introduced to reduce the number of clusters for which the inner loop computation needs to be performed.  Based on the property of stochastic \ac{em} \cite{nielsen2000stochastic}, we prove that convergence (to a local minimum) is still guaranteed.

 semicolon-separated keywords here!


\section{Preliminaries}
Notations: Let $S$ be a finite discrete set, $\dist(S)$ is the set of distributions over the support $S$.

We assume that the system (including the environment) is modeled as a \ac{mdp} $(S,A,P,r,\gamma, \rho_0)$ where $S$ is the set of states, $A$ is the set of actions, $P: S\times A\rightarrow \dist(S)$ is the transition probability, and $P(s'| s,a)$ is the probability of reaching $s'$ upon taking action $a$ at the state $s$, $r:S\times A\rightarrow \reals$ is the reward function, and $\gamma$ is the discount factor, and $\rho_0\in \dist(S)$ is the initial probability distribution of states. A memoryless policy is a mapping $\pi: S \rightarrow \dist(A)$ where $\pi(a| s)$ is a distribution of actions in $A$. From an \ac{mdp} $M$, the policy $\pi$ induces a Markov chain $M^\pi = (S,A, P^\pi, \rho_0)$ where $P^\pi :S\rightarrow \dist(S)$ where $P^\pi (s' | s) = \sum_{a\in A}P(s' | s,a)\pi(a| s)$.
\sloppy
The value of $\pi$ is the expected discounted return of carrying out the policy $\pi$, defined as $V^\pi = \expect\left[ \sum_{t=0}^\infty \gamma^t r(S_t,A_t)| \rho_0, \pi \right]$ where $S_t$ and $A_t$ are the $t$-th state and action in the Markov chain $M^\pi$.
It can be computed by $V^\pi(s) = \sum_{a\in A} r(s, \pi(a| s)) \pi(a| s) + \gamma\sum_{s'\in S}\sum_{a\in A} P(s'| s, a) \pi(a| s) V^\pi(s').$ 
Let the set of memoryless policies be denoted as $\Pi$.  Given an \ac{mdp}, the planning objective is to compute an optimal policy $\pi^\ast$ that maximizes the value: $V^\ast(s) = V^{\pi^\ast}(s) = \max_{\pi \in \Pi} V^\pi(s)$ for all $s\in S$.

\subsection{Maximum entropy inverse reinforcement learning}

The inverse reinforcement learning problem in \ac{mdp}s aims to find a reward function $r: S\times A\rightarrow \reals$ such that the distribution of state-action sequences under a (near-)optimal policy match the demonstrated behaviors.  It is assumed that the reward function $r$ is given as a linear combination of $k$ features $\{\phi_i, i=1, \ldots, k\}$ such that $\forall (s,a) \in S \times A: r_\theta(s,a) = \theta\cdot\phi(s,a)$ where $\theta \in \reals^k$ is a vector of unknown parameters and $\phi(s,a)= [\phi_i(s,a)]_{i=1}^k$ is a vector of features, and the dot $\cdot$ is the inner product.  Given a trajectory $\tau$ in the form of a sequence $((s_0,a_0), (s_1,a_1),(s_2,a_2), \ldots, (s_T,a_T))$ of states $s_t\in S$ and actions $a_t\in A$, for $t=0,\ldots, N$, the discounted return of this trajectory can be written as
\begin{equation}
	\begin{split}
		r_\theta (\tau) = \sum_{t=0}^{T-1} \gamma^t r(s_t,a_t ) \approx \theta\cdot\phi(\tau),
	\end{split}
\end{equation} where $\phi(\tau) = \sum_{t=0}^{T-1}\gamma^t \phi(s_t,a_t) $ is the discounted feature vector counts along trajectory $\tau$ and $\theta\cdot\phi(\tau)$ approximates  the discounted return.

One of the well-known solutions to Maximum Entropy IRL problem~\cite{ziebart2008maximum} proposes to find the policy, which best represents demonstrated behaviors with the highest entropy, subject to the constraint of matching feature expectations to the distribution of demonstrated behaviors.

Let $\traj$ be the set of trajectories generated by the Markov chain $M^\pi$ where $\pi$ is the maximum entropy policy, probability of a trajectory $\zeta$ under this policy is given by,
$P(\zeta | \theta) = \frac{\exp(r_\theta(\zeta))}{\sum_{\tau \in \traj}\exp(r_\theta(\tau))}$. 
Let $\mathcal{D}$ be the set of demonstrations. The log likelihood
of the demonstrations and its derivative are given by
\begin{align*}
	\mathcal{L}(\theta) &= \frac{1}{n} \sum_{\tau^i \in \mathcal{D}} \ln P(\tau^i|\theta)\\
	\frac{\partial \mathcal{L}(\theta)}{\partial \theta}
	&= \frac{1}{n} \sum_{\tau^i \in \mathcal{D}} \frac{\partial r_\theta}{\partial \theta}(\tau^i) - \sum_{s \in S,a \in A} D(s,a|\theta)\frac{\partial r_\theta}{\partial \theta}(s,a)   \nonumber
\end{align*} where $D(s,a|\theta)$ is the state-action visitation
frequency in the maximum entropy policy induced Markov chain and can
be obtained using soft-value iteration \cite{ziebart2010modeling} with
respect to the current reward parameters $\theta$. 
Let $\phi(\tau)$ be the discounted
feature expectation along the trajectory $\tau$ \ie,
\begin{equation}
	\label{eq:feature_expectation} \phi(\tau^i) = \sum_{t=0}^{N-1} \gamma^t\phi(s_t,a_t).\end{equation}  The gradient of log
likelihood function becomes
\begin{align}
	\frac{\partial \mathcal{L}(\theta)}{\partial \theta} &= \frac{1}{n} \sum_{\tau^i \in \mathcal{D}} \phi(\tau^i) - \sum_{s \in S,a \in A} D(s,a|\theta)\phi(s,a)  \nonumber \\ 
	\label{linear_gradient}
	&= \bar{\phi}_\mathcal{D} - \bar{\phi}_\mathcal{\pi}
\end{align} 
where $\bar{\phi}_\mathcal{D}$ is the feature expectation of the
expert demonstrations and $\bar{\phi}_\pi$ is the feature expectation
of the steady state distribution under the maximum entropy policy.

The MaxEnt-IRL initializes the parameters $\theta$ randomly and uses gradient descent to maximize the objective $\mathcal{L}(\theta)$. In linear reward setting, we can interpret the gradient in \eqref{linear_gradient} as the difference between the feature expectation of expert and that of the maximum entropy policy under the current reward function approximation.

The following problem is studied:

\begin{problem}
	\label{prob:main} We are given an \ac{mdp} without the reward
	function, that is, the tuple $(S,A,P,\gamma)$, and the set of
	demonstrations $\mathcal{D}$ from the
	expert(s). 
	We assume that each of these demonstrations can arise probabilistically from several different behaviors $\{c_j| j=1,\ldots, m\}$ where $m$ may be unknown. 
	How to simultaneously cluster behavior and inverse learning the reward function of each cluster?
\end{problem}

We first present an \ac{em}-based approach for \ac{irl} with a known number of clusters and then show how to solve the problem when the number of clusters is unknown a priori. 

\section{Parametric Behavior Clustering IRL }
\label{parametric}

In this section, we assume the number of clusters is given and develop an algorithm to solve Problem~\ref{prob:main}.  Let $\mathcal{C} = \{c_j | 1\le j \le m\}$ be the set of different behaviors. We assume that there exists a prior probability distribution $\Psi$ over $\mathcal{C}$, \ie, $\Psi(c_j)$ is the probability that a randomly selected demonstration belongs to the cluster $c_j$. Each behavior cluster $c_j$ has a unique parameter $\theta_j$ for its reward function, \ie, $r_{\theta_j} (s,a)= \theta_j\cdot \phi(s,a)$ where it is assumed that the feature vectors among different behaviors are the same $ \phi= \left[\phi_j \right]_{j=1}^n$. 

Let $\Theta = \{\theta_j \mid 1\le j \le m\}$ be the collection of reward parameters, one for each behavior. 
Formally, the data is generated from the distribution $P(\tau| \Theta,\Psi)$ and takes the form of demonstrations $\mathcal{D} =\{\tau^j, j=1, \ldots, n\}$. We formulate the problem of finding the underlying behavior as an \ac{em} \cite{College02theexpectation} problem and treat the demonstrations $\mathcal{D}$ as the observed data and the class assignment $c_j$ (comes from $j$-th behavior) for each demonstration as the missing data. Our objective is to maximize the probability of obtaining these demonstrations, \ie,
$P(\mathcal{D} | \Theta,\Psi)$.  The objective function of our \ac{em} 
algorithm is given by,

\begin{align}
	\max_{\Theta,\Psi} \ln P(\mathcal{D}|\Theta,\Psi)
\end{align}

Assuming that each demonstration was generated independent of the other, we can rewrite the objective as 
\begin{align}
	\label{exact}
	& \max_{\Theta,\Psi} \sum_{i=1}^n \ln( P(\tau^i|\Theta,\Psi)) \\
	\label{exp}
	&\max_{\Theta,\Psi}\sum_{i=1}^n \sum_{j=1}^m P(c_j|\tau^i, \Theta, \Psi) \ln(P(c_j,\tau^i|\Theta,\Psi))
\end{align}
We can get \eqref{exp} from \eqref{exact} using the EM formulation of the objective function  (also shown in appendix)  \cite{dempster1977maximum}. To compute the posterior $P(c_{j}|\tau^i, \Theta,\Psi)$ of the cluster distribution we use the Baye's rule and define 

\begin{align}
	\label{def_beta}
	\beta_{ij}^{(t)} = \frac{P(\tau^i| c_j,\Theta^{(t)},\Psi^{(t)})\Psi^{(t)}(c_{j})}{\sum_{k=1}^m P(\tau^i| c_k,\Theta^{(t)},\Psi^{(t)})\Psi^{(t)}(c_{k})}
\end{align} where $\beta_{ij}^{(t)}$ is the probability that
demonstration $\tau^i$ comes from the cluster $c_j$, \ie,
$P(c_j|\tau^i)$ at the $t$-th iteration of the EM algorithm. Let
$\pi_j$ be the MaxEnt policy given the reward function of the
$j$-th cluster. We compute $P(\tau^i|c_{j},\Theta,\Psi)$ using 
\begin{align*}
	P(\tau^i|c_{j},\Theta,\Psi) &= P(\tau^i|\theta_j)\\
	&= \prod_t \pi_j(a_t|s_t) P(s_{t+1}|a_t,s_t)
\end{align*} where $(s_t,a_t )\in \tau^{i}$ for $t\ge 0$ are the state-action pairs in the trajectory $\tau^{i}$.

Given no knowledge about the number of demonstrations that come from each behavior, we can initialize the prior $\Psi$ to the maximum entropy solution (uniform distribution), \ie, $\Psi^{(0)}(c_{j}) = P(c_{j}|\Theta,\Psi) = \frac{1}{m}$ for all $ j$ for the $0$-th iteration.

At the $t$-th iteration, $t\ge 1$, the objective function can now be written using $\beta_{ij}^{(t)}$ as,
\[
\mathcal{L}^{(t)}(\Theta,\Psi)
=\sum_{i=1}^n \sum_{j=1}^m \beta_{ij}^{(t)}  \ln\big(P(\tau^i| c_{j},\Theta,\Psi)\Psi(c_{j})\big)
\]

We update the parameters by performing the M-step.
\begin{align*}
	&\Theta^{(t+1)},\Psi^{(t+1)} \\
	= &\arg \max_{\Theta,\Psi} \sum_{i=1}^n \sum_{j=1}^m \beta_{ij}^{(t)} \ln\big(P(\tau^i| c_j,\Theta,\Psi)\Psi(c_{j})\big) \\
	&\text{s.t. } \sum_j \Psi(c_{j}) = 1
\end{align*}
The objective can be written as, 
\begin{align}
	\label{final_objective}
	\max_{\Theta,\Psi} \sum_{i=1}^n \sum_{j=1}^m
	\beta_{ij}^{(t)} &\Big[\ln\big(P(\tau^i|
	c_j,\Theta,\Psi)\big)  + \ln\big(\Psi(c_{j})\big)\Big]
\end{align}

The first term in the summation is conditionally independent of $\Psi$ given $c_j$ and the second term is independent of $\theta$. Hence we can maximize the terms separately as two independent problems. Note that the constraint $\sum_j \Psi(c_j) = 1$ applies to the second term only.  We now have two independent problems in M-step: 1) Maximizing the likelihood of demonstrations and 2) Maximizing the likelihood of the prior.

\subsubsection{Problem 1:}Maximizing the likelihood of the
demonstration: 
\begin{align}
	\max_{\Theta,\Psi} \sum_{i=1}^n \sum_{j=1}^m \beta_{ij}^{(t)}
	\Big[\ln\big(P(\tau^i | c_j,\Theta,\Psi)\big)\Big]
\end{align} 

As this objective function is independent of $\Psi$, we can change the order of summation and maximize the likelihood of the demonstration individually for each cluster $c_j$ given the probability $\beta_{ij}^{(t)}$ of the demonstration $\tau^j$ in the $j$-th cluster, for all demonstrations $\tau^j\in \mathcal{D}$.
\begin{align}
	\label{opt_problem}
	\theta_j^{(t+1)} = &  \max_{\theta_j}\sum_{i=1}^n\beta_{ij}^{(t)} \ln\big(P(\tau^i| \theta_j)\big)
\end{align} 
which is the same as the log likelihood of demonstrations under a single reward function \cite{ziebart2008maximum} except that each likelihood is now weighed by $\beta_{ij}^{(t)}$. The gradient of $\mathcal{L}_m(\theta_j)$ is now given by, 
\begin{align}
	\nabla_{\theta_j}\mathcal{L}_m(\theta_j) &= \sum_{i=1}^n
	\beta_{ij}^{(t)}\phi^i
	-
	\sum_{s\in S,a\in A}\beta_{ij}^{(t)}D(s,a
	| \theta_j)\phi(s,a) \nonumber \\
	&= \bar{\phi}_{\mathcal{D},j} - \bar{\phi}_{\pi_j}
\end{align} where $\phi^i$ is the feature expectation of the $i$-th demonstration as in \eqref{eq:feature_expectation}, $D(s,a |\theta_j)$ is the state-action visitation frequency using the maximum entropy policy given reward function $\theta_j\cdot \phi(s,a)$, the term $\bar{\phi}_{\mathcal{D},j}$ is the \emph{weighted} feature expectation of the demonstrations, and $\bar{\phi}_{\pi_j}$ is the \emph{weighted} feature expectation of the maximum entropy policy $\pi_j$ given the reward $\theta_j \cdot \phi(s,a)$.

Intuitively, if the probability of a demonstration $i$ coming from behavior $j$ is very low, \ie, $\beta_{ij}^{(t)}$ is close to zero, then the contribution of the gradient from $i$-th demonstration to $j$-th reward parameters is almost zero. 


\subsubsection{Problem 2:}
Maximizing the likelihood of the prior
\begin{align*}
	&\Psi^{(t+1)} = \arg \max_{\Psi} \sum_{i=1}^n \sum_{j=1}^m \beta_{ij}^{(t)} \ln\big(\Psi(c_{j})\big) 
	\text{s.t. } \sum_{k=1}^m \Psi(c_{k}) = 1
\end{align*} 

The Lagrangian of Problem 2 is given by,
\begin{align*}
	\Gamma &= \sum_{i=1}^n \sum_{j=1}^m \Big[\beta_{ij}^{(t)} \ln\big(\Psi(c_{j})\big)\Big] + \lambda\Big[\sum_{k=1}^m \Psi(c_{k}) - 1\Big]
	\text{s.t. } \lambda \ne 0.
\end{align*} 
Setting the derivative of the Lagrangian to zero, we get, 
\begin{align}
	\frac{\partial\Gamma}{\partial\Psi(c_{j})} &=
	\frac{1}{\Psi(c_{j})}\sum_{i=1}^n
	\beta_{ij} ^{(t)}+
	\lambda = 0 \nonumber 
	\implies \Psi(c_{j}) \propto \sum_{i=1}^n \beta_{ij}^{(t)} \nonumber \\ 
	\label{eq:norm}
	\Psi^{(t+1)}(c_{j}) &= \frac{\sum_{i=1}^n \beta_{ij}^{(t)}}{\sum_{j=1}^m \sum_{i=1}^n\beta^{(t)}_{ij}} =\frac{\sum_{i=1}^n \beta^{(t)}_{ij}}{n}
\end{align} 
\eqref{eq:norm} comes from the constraint that the probabilities $\Psi(\cdot)$ must sum to one and $\sum_{j=1}^m \beta_{ij}^{(t)} =  1$. 

In this section, we developed an \ac{em}-approach that iteratively
updates the prior probability distribution $\Psi$ over clusters and
the reward functions of all clusters given the number of clusters is
known. Next, we provide a solution to Problem~\ref{prob:main} given
the number of clusters is unknown.

\section{Non-parametric Behavior Clustering IRL}

In this section, we present the Non-parametric Behavior Clustering \ac{irl}, \ie, Non-parametric BCIRL, that learns multiple reward functions when the number $m$ of clusters is unknown. We assume each trajectory is generated by an agent with a fixed reward function using the corresponding maximum entropy policy. Similar to \cite{choi2012nonparametric}, we use the \ac{crp} to learn multiple behavior clusters from data, while the reward function in each cluster is learned through MaxEnt \ac{irl} instead of Bayesian \ac{irl}. However, a naive alternation between nonparametric clustering and MaxEnt \ac{irl} can be computational intensive as solving \ac{irl} involves iteratively solving the forward reinforcement learning problem with each gradient-based reward function update. Besides, the number of \ac{irl} problems to be solved in each iteration increases with the number of clusters generated. In the worst case, nonparametric clustering can generate multiple clusters, each of which contains only a few demonstrations. To this end, we propose two methods to improve the convergence of behavior clustering and learning of multiple reward functions: First, in the iterative procedure, our algorithm does not solve the complete \ac{irl} problem for each cluster. Rather, it alternates between one (or multi-) step update in reward functions and clustering. Thus, it will reduce the computation comparing to the case when one needs to solve multiple \ac{irl} during each iteration. Second, we introduce a resampling technique to effectively reduce the number of clusters that need to be considered during each iteration. We show that the convergence can still be guaranteed based on the principle of \ac{em} and resampling methods. 


\paragraph*{Preliminary:  \ac{crp}}
The \ac{crp} model is a sequential construction of partitions used to define a probability distribution over the space of all possible partitions \cite{aldous1985exchangeability}. The concentration parameter $\alpha$ of \ac{crp} controls the probability that a new data-point starts a new cluster.  Using the \ac{crp} prior, the probability of a demonstration being associated with an existing cluster is proportional to the probability density of demonstrations in that cluster, and that of a new cluster is proportional to the concentration parameter $\alpha$ in \eqref{crp_eq}.
\begin{align} 
	\label{crp_eq}
	c_j | c_1, c_2 ,..., c_n = 
	\begin{cases}
		c_k  &\text{w.p. } \frac{\sum_{i} \beta_{ik}}{n+\alpha} \\
		\text{new cluster } &\text{w.p. } \frac{\alpha}{n+\alpha}
	\end{cases}
\end{align}

The posterior $P(c_j|\tau^i)$ is given by
\begin{equation}
	\label{posterior_prior}
	\beta_{ij} \triangleq P(c_j|\tau^i) = \frac{1}{\eta}P(\tau^i|c_j)P(c_j)
\end{equation} where $\eta$ is the normalizing constant. This posterior always has a non-zero probability of starting a new cluster at every iteration. At one point, we would need to solve the reinforcement learning problem as many times as the number of demonstrations which is computationally demanding. To alleviate this problem, we use stochastic EM \cite{nielsen2000stochastic} and perform \emph{weighted re-sampling/bootstrapping} from the posterior to get rid of residual probability masses in new clusters. Moreover, instead of completely solving \ac{irl} problems for all clusters during each iteration in the inner loop, we only take one step in the gradient ascent method for solving the \ac{irl} problem. We show that the algorithm is guaranteed to converge (to a local minimum due to the \ac{em} method) in Appendix.

\paragraph*{Algorithm}
Algorithm~\ref{EMIRL} concludes our Nonparametric BCIRL that includes the outer loop with nonparametric clustering and the inner loop with stochastic \ac{em} for inverse learning multiple reward functions from demonstrations. The weighted resampling is the \texttt{bootStrap} function in line 16 of Algorithm.\ref{EMIRL}.  This function is analogous to resampling in particle filter to make the best use of the limited representation power of the particles. This also reduces high time complexity by redistributing the particles at high probability masses and eliminating particles at low probable regions. This re-sampling does not affect the point of convergence \cite{nielsen2000stochastic} since in an expectation over several iterations we would be sampling from the original distribution ($\beta'_i$ in the algorithm). Lines 15 and 16 in Algorithm.\ref{EMIRL} can be interpreted as removing the $i$-th demonstration from its existing cluster assignment and reassigning them.

\begin{algorithm}[!h]
	\caption{Non-parametric BCIRL(\ac{mdp} $(S, A, P, \gamma)$, $\mathcal{D}$, $\alpha, RL$)}\label{EMIRL}
	\begin{algorithmic}[1]
		\INPUT\tikzmark{a} 
		\Statex $\mathcal{D} \to$ The dataset of demonstrations
		\Statex $\alpha \to$ The parameter of CRP
		\Statex $RL$ Reinforcement Learning algorithm to solve the MaxEnt policy 
		\tikzmark{b}
		\OUTPUT\tikzmark{a} 
		\Statex The number $m$ of clusters, a distribution of clusters for each demonstration, \ie, $\{P(c_j|\tau^i)\mid i=1,\ldots, n, j=1,\ldots, m \}$, and the reward function $r_j$ of each cluster
		\tikzmark{b}
		
		\State $n \gets \text{Number of demonstrations in } \mathcal{D}$
		\State $\texttt{nc} \gets \texttt{[]}$  \Comment{Number of demonstrations in each cluster}
		\State $P \gets$ The transition probability of the MDP 
		\State $\beta_{i}\gets 0  \ \ \forall i=1:n$
		\State \texttt{$ss \gets$ startStates($\mathcal{D}$)}
		\While {$True$}
		\State \texttt{m $\gets$ len(nc)}
		\State $\theta_{(m+1)} \gets \text{Random Sample from prior}$
		\label{add_newcluster}
		\State \texttt{p $\gets$ normalize(merge(nc, [$\alpha$]))}
		\For{$i = 1:n$} 
		\For{$j = 1:m$} 
		\State $\pi_j \gets RL(\theta_j)$
		\State $\beta'_{ij} \gets \texttt{p[j]} \cdot \prod_t \pi_j(a_t|s_t)P(s_{t+1}|s_t,a_t)$ where $s_t,a_t \in \tau^i$
		\EndFor
		\State $\beta'_{ij} \gets \frac{\beta'_{ij}}{\sum_j\beta'_{ij}}$ \Comment{normalize $\beta'_{ij}$}
		\State \texttt{nc = nc - $\beta_i$} \Comment{element wise subtraction}
		\State $\beta_i \gets$ \texttt{bootStrap($\beta'_i$)} \Comment{weighted re-sampling from the distribution $\beta_i'$}
		\State \texttt{nc = nc + $\beta_i$} \Comment{reseat $i$-th demonstration according to new distribution\footnotemark}
		\State \texttt{nc.sparsify()} \Comment{remove zero entries}
		
		\EndFor
		
		\For{$j = 1:m$} 
		\State \texttt{$D_j \gets$ visitationFromDemos($\mathcal{D}$, $\beta_j$)}
		\State \texttt{$D_{\pi_j} \gets$ visitationFromPolicy($\pi_j$,$ss$, $\beta_j$)}
		\State $\bar{\phi}_\mathcal{D} \gets \sum_{s,a} D_j(s,a)\phi(s,a)$
		\State $\bar{\phi}_{\pi_j} \gets \sum_{s,a} D_{\pi_j}(s,a)\phi(s,a)$
		\State $\theta_j \gets \theta_j - \alpha(\bar{\phi}_\mathcal{D} - \bar{\phi}_{\pi_j})$
		\State $\Psi(c_j) \gets \frac{\sum_i\beta_i}{n}$
		\If {$||(\bar{\phi}_\mathcal{D} - \bar{\phi}_{\pi_j})||_\infty < threshold$} \Comment{if the gradient $\to 0$}
		\State \Return $\theta$
		\EndIf
		\EndFor
		\EndWhile
	\end{algorithmic}
\end{algorithm}
\footnotetext{This is element-wise addition. Might increase the size of the list when size of $\beta_i$ is greater than that of $nc$}

The non-parametric BCIRL with CRP not only solves the problem of learning the number of clusters based on the likelihood of the demonstrations but also allows EM to escape local minima created in the beginning. As we will see in the results section, the parametric BCIRL often gets stuck in local minima. Non-parametric BCIRL detects the presence of a new cluster using non-parametric clustering (CRP) and thus saves the efforts of deciding a problem specific threshold variance beyond which we should create a new cluster.
Though the parameter $\alpha$ in Algorithm.\ref{EMIRL} can be seen as a means to set this threshold, it is a robust parameter \ie the algorithm is stable for a range of values of $\alpha$. 

\section{Experiments and Results} In this section, we demonstrate the correctness and efficiency of the proposed algorithms with experiments. The experiments were implemented on a laptop computer with Intel Core™ i7-5500U CPU and 8 GB RAM.  
\subsection{Simple Gridworld Task}
\begin{figure}
	\begin{center}
		\includegraphics[width=1\linewidth]{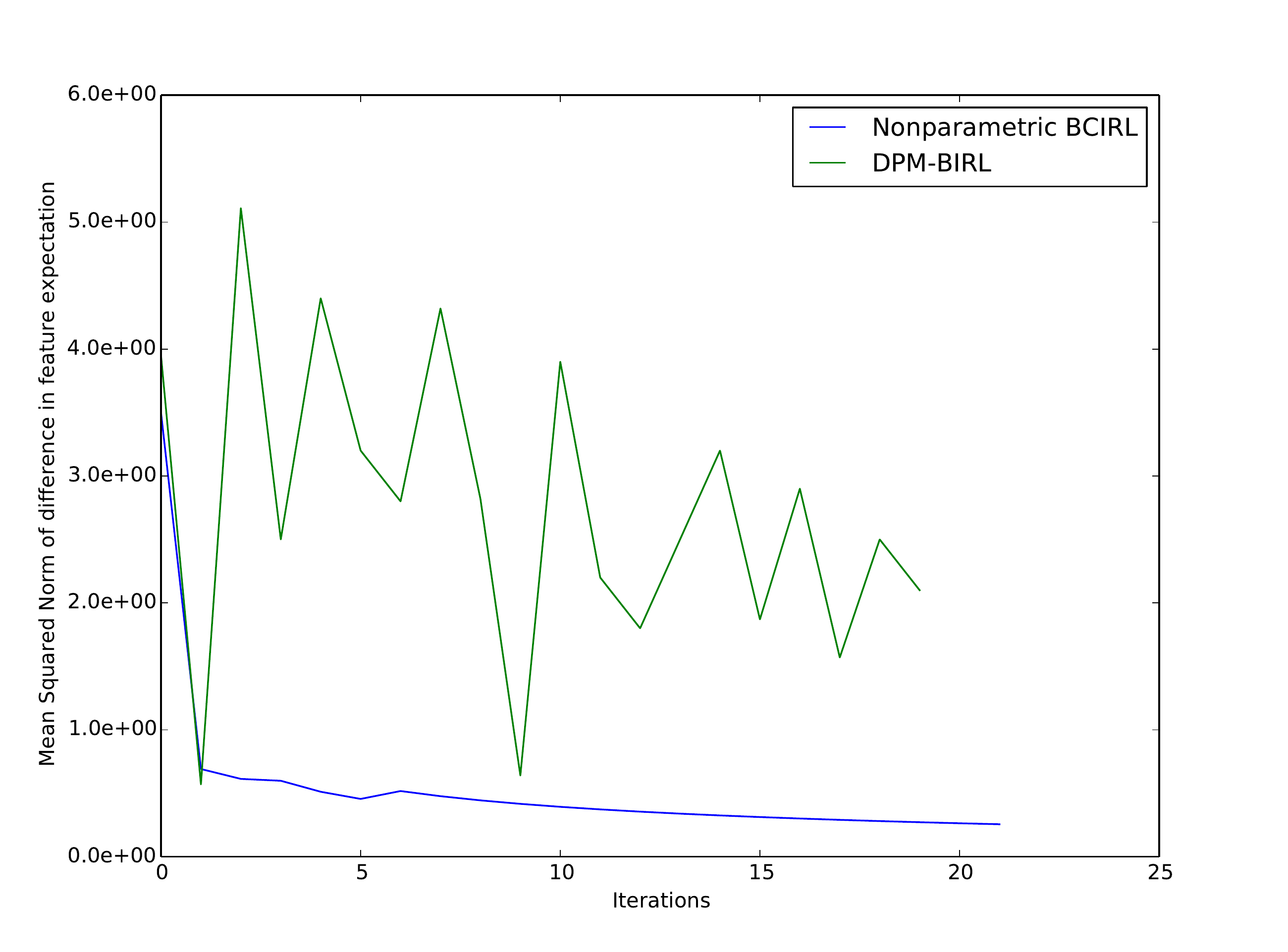}
		\caption{Comparison of difference in learned and the expert behavior using the difference in their feature expectations.}
		\label{fig:comp}
	\end{center}
\end{figure}

We first compare our algorithm with the Nonparametric Bayesian IRL \cite{choi2012nonparametric} on a gridworld of dimension $8\times8$  where each cell corresponds to the state. At any givent time, the agent occupies a single cell and can take the move north, south, east, or west, but with a probability of 0.2, it fails and moves in a random direction. The cells are partitioned into macro-cells of dimension $2\times 2$ which are binary indicator features. We generated trajectories using 3 different reward functions. The reward functions are generated as follows: for each macro-cell, with $80\%$ probability, we set the reward to zero and otherwise we set it to a random number from $-1$ to $1$. We generated 1-trajectory for each reward function and compared the performance of the two algorithms. As a performance metric, we find the mean-squared norm of the difference in the feature expectation of the learned behavior and the expert behavior. Since the method in \cite{choi2012nonparametric} involves solving the IRL problem using Metropolis-Hastings sampling, the convergence of the proposed metric is not monotonic for this problem. Fig.\ref{fig:comp} shows the above comparison. To detect the correct number of underlying clusters, it took on an average (over $100$ runs) $13$ iterations in hundred runs of the Nonparametric Bayesian IRL while it took only $7$ iterations using our Nonparametric BCIRL.

\subsection{Gridworld Driving Task}
\begin{figure}
	\begin{center}
		\begin{minipage}{0.5\textwidth}
			\centering
			\includegraphics[width=1\textwidth]{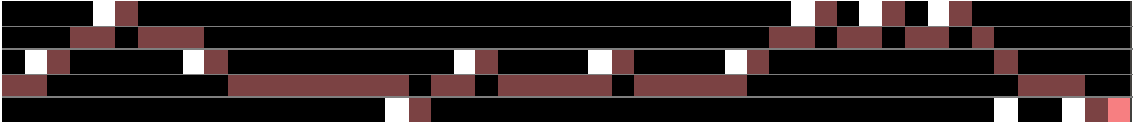} 
			\caption*{(a) Aggressive demonstration}
		\end{minipage}\hfill
		\begin{minipage}{0.5\textwidth}
			\centering
			\includegraphics[width=1\textwidth]{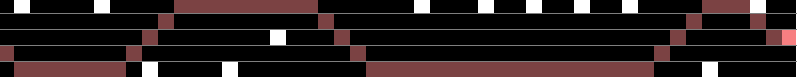} 
			\caption*{(b) Evasive demonstration}
		\end{minipage}\hfill
		\begin{minipage}{0.5\textwidth}
			\centering
			\includegraphics[width=1\textwidth]{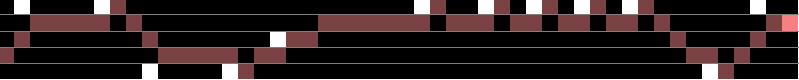} 
			\caption*{(c) Aggressive learned behavior using BCIRL}
		\end{minipage}\hfill
		\begin{minipage}{0.5\textwidth}
			\centering
			\includegraphics[width=1\textwidth]{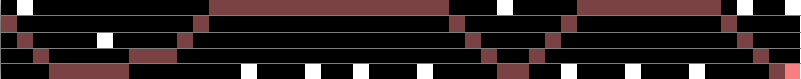} 
			\caption*{(d) Evasive learned behavior using BCIRL}
		\end{minipage}\hfill
		\begin{minipage}{0.5\textwidth}
			\centering
			\includegraphics[width=1\textwidth]{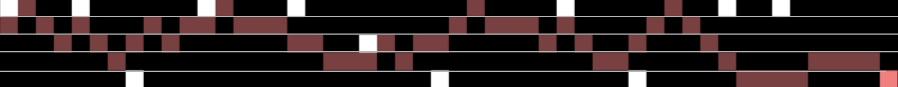} 
			\caption*{(e) Single learned behavior using MaxEnt IRL (neither evasive nor aggressive)}
		\end{minipage}\hfill
	\end{center}
	\caption{The grid-world simulating driving behaviors on a five lane road. The left of each of the figure is the start position, and the right is the end position (all cars moving from left to right). Dark pink (maroon) shows the path taken by the driving behavior and the lighter pink (in far right) is the final position of the agent. 
	}
	\label{fig:grid_race}
\end{figure}

In this experiment, we simulate different driving behaviors in a
discrete gridworld. The gridworld can be considered as a
discretization of the continuous state space. In
Fig.~\ref{fig:grid_race}, the white cells are occupied by other
vehicles, which have randomly initialized positions and move at a
constant speed. The sequence of maroon cells is a path taken by an
agent of interest, who moves faster than the others. Two different
driving styles, aggressive and evasive, are demonstrated. An
aggressive driver tries to cut in front of other cars after overtaking
them while an evasive driver tries to stay as far away from the other
cars as possible. To generate demonstrations, we hard code their
policies which map states to distributions over the possible
actions (right, left, hard-right, hard-left, forward and break),
depending on the relative pose of the agent and other cars. The
executed action is successful with probability $0.95$. Otherwise, the
agent stays put with probability $0.05$. The aggressive driver has a
very high probability to cut in front of other cars while the evasive
driver has a high probability to move away from other cars while
moving forward.  For this task, we use a state action indicator
feature for all actions when the agent is far away from the next car,
in the vicinity and while overtaking the next car. Also, we use an additional
feature---horizontal distance to goal (the right end in Fig.\ref{fig:grid_race}).

\begin{figure}
	\begin{center}
		\includegraphics[width=1\linewidth]{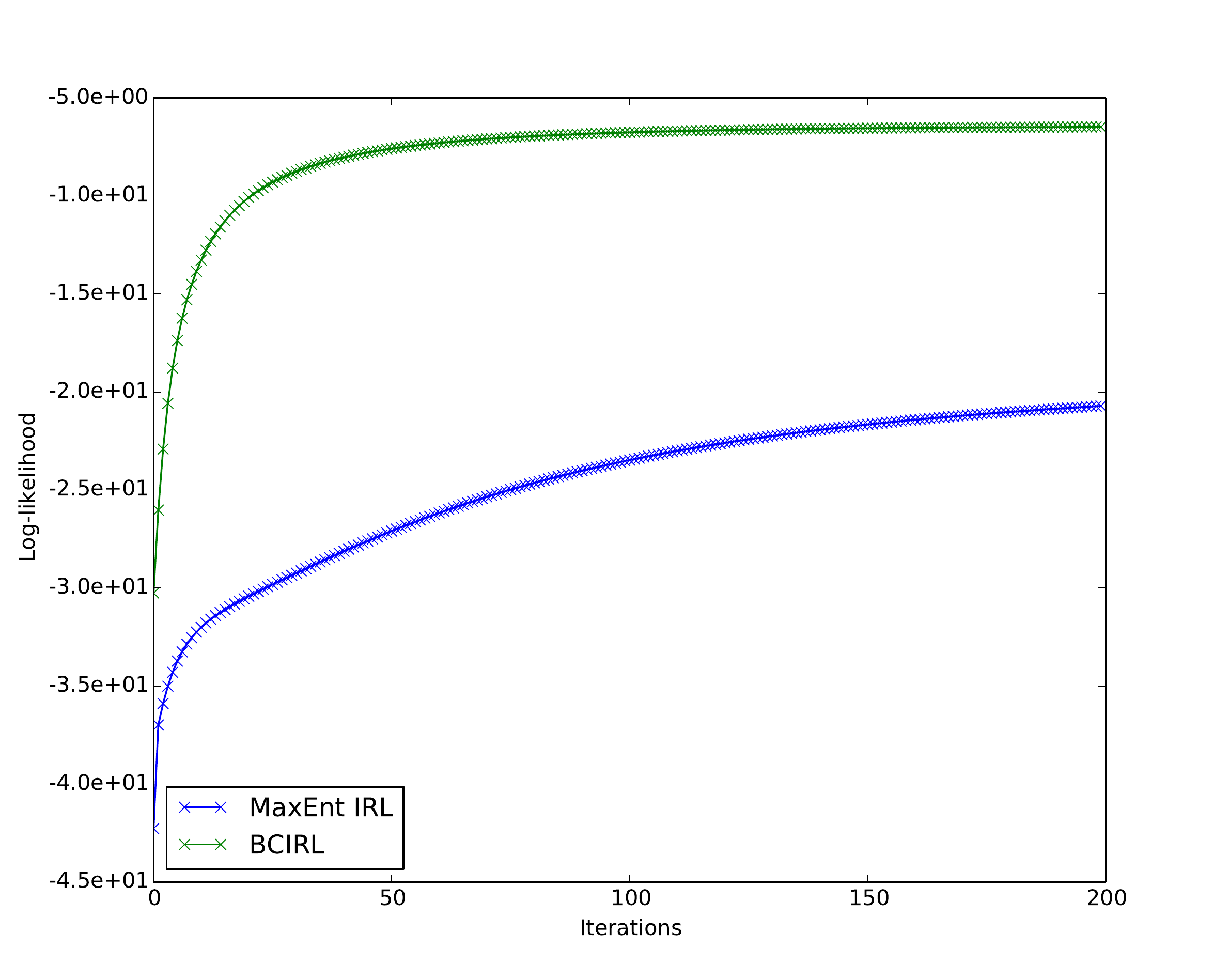}
		\caption{The value of the objective function vs. iterations using Nonparametric Behavior Clustering IRL and MaxEnt IRL.}
		\label{fig:bcirl_obj}
	\end{center}
\end{figure}

We ran the Nonparametric BCIRL on a dataset with $25$ aggressive and $25$ evasive demonstrations and compared it to MaxEnt IRL. In Fig.\ref{fig:bcirl_obj}, we compare the value of the objective, \ie,
log likelihood of demonstrations given reward function(s), using BCIRL
vs. vanilla MaxEnt IRL. The MaxEnt IRL started with a likelihood of
$10^{-14}$ and converged at $10^{-11}$ while BCIRL converged at
$1.5 \cdot 10^{-3}$.  In fact, MaxEnt IRL converged to a policy that explains both the behaviors simultaneously and hence the low likelihood of observing the demonstrations under this policy. The behavior learned by MaxEnt IRL can be seen in Fig.~\ref{fig:grid_race}(e). We also noted that the parametric BCIRL with the correct number of parameters ($2$ in our case) often converged to a local minima. That is, it went on to classify all the demonstrations into one cluster and learned the MaEnt IRL solution for that cluster. This, however, does not happen for the nonparametric version since, in the early phase of the clustering process, Dirichlet Process is known to create new clusters to escape these local minima \cite{murphy2012machine}. 

We expect to see higher time consumed per iteration in case of BCIRL since it solves as many RL problems as the number of clusters in each iteration while the MaxEnt IRL invariably solves only one RL problem per iteration. Using BCIRL, the time per iteration starts with 1.4 seconds (on average) in the first 25 iterations and becomes 0.91 seconds per iteration (on average) in the last 25 iterations. The average time per iteration is 0.4 sec for MaxEnt \ac{irl}. It is reasonable since initially, we observed a high number of
clusters which then gradually reduced to only two clusters (aggressive
and evasive behaviors) in less than $20$ iterations. The rest of the
iterations learn the reward functions more accurately.

We also observed that the algorithm is very robust to the parameter
$\alpha$ of the CRP. BCIRL, with a low value of $\alpha$, tends to get stuck in
the local minima in the initial phases of convergence while a very
large value detects the correct number of clusters early on but takes
a long time to converge. For all the tasks, we use
$\alpha = 3$ (and values ranging between $3$ and $10$ gave similar
results).

\subsection{Simulator Driving Task}

We collected demonstrations from autonomous robot car driving using the Gazebo simulator in ROS \cite{quigley2009ros}.  We used potential fields \cite{goerzen2010survey}, a heuristic control, to generate our data set to show that the method extends well even for sub-optimal expert demonstrations with different behaviors. 

In potential field approach, the agent is considered as a point under the influence of the fields produced by goals and obstacles. A repulsive potential field is generated by an obstacle to encourage the agent to avoid obstacles, whereas an attractive one is generated by a goal to attract the agent to reach the goal. The force generated by the potential field is applied to the agent to enable reach-avoid behaviors.

We use a linear attractive potential (goal) at the far end of the course. Each car (other than the agent) and either side of the road is associated with a repulsive Gaussian potential. We generate two behaviors similar to the grid world driving task using the potential field in a continuous environment \footnote{The video can be found at \url{https://goo.gl/vfn4gB}}.

\begin{figure}[H]
	\centering
	\begin{minipage}{0.11\textwidth}
		\centering
		\includegraphics[width=1\textwidth]{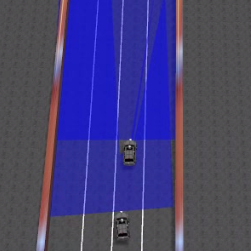} 
	\end{minipage}\hfill
	\begin{minipage}{0.11\textwidth}
		\centering
		\includegraphics[width=1\textwidth]{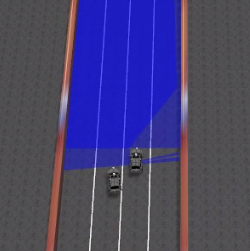} 
	\end{minipage}\hfill
	\begin{minipage}{0.11\textwidth}
		\centering
		\includegraphics[width=1\textwidth]{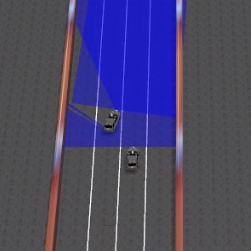} 
	\end{minipage}\hfill
	\begin{minipage}{0.11\textwidth}
		\centering
		\includegraphics[width=1\textwidth]{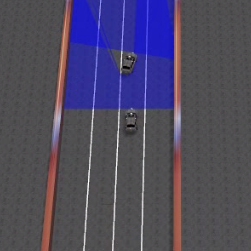} 
	\end{minipage}\hfill
	\caption{Snapshots of the video to show the aggressive driving behavior in Gazebo simulator in the order from left to right.}
	\label{fig:agg_overtake}
\end{figure}

\begin{figure}[H]
	\centering
	\begin{minipage}{0.11\textwidth}
		\centering
		\includegraphics[width=1\textwidth]{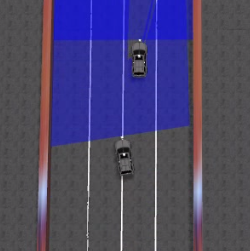} 
	\end{minipage}\hfill
	\begin{minipage}{0.11\textwidth}
		\centering
		\includegraphics[width=1\textwidth]{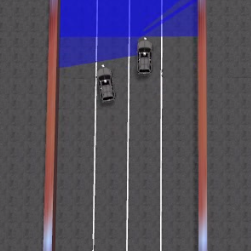} 
	\end{minipage}\hfill
	\begin{minipage}{0.11\textwidth}
		\centering
		\includegraphics[width=1\textwidth]{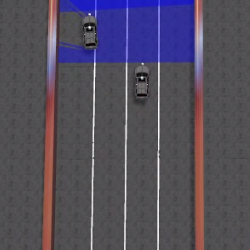} 
	\end{minipage}\hfill
	\begin{minipage}{0.11\textwidth}
		\centering
		\includegraphics[width=1\textwidth]{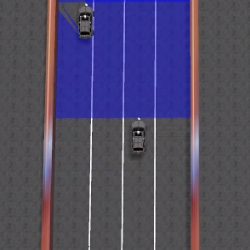} 
	\end{minipage}\hfill
	\caption{Snapshots of the video to show the evasive driving behavior in Gazebo simulator in the order from left to right.}
	\label{fig:ev_overtake}
\end{figure}

\begin{enumerate}
	\item \textbf{Aggressive:} We use a weaker repulsive potential associated with other cars to allow close interactions. Also, each of the other cars is associated with an attractive potential in front of them to allow the aggressive agent to cut in front of other cars. Aggressive behavior is shown in Fig.\ref{fig:agg_overtake}.
	\item \textbf{Evasive:} We use one stronger repulsive potential centered at other cars to make our agent be evasive. Evasive behavior is shown in Fig.\ref{fig:ev_overtake}.
\end{enumerate}

We generated $25$ demonstrations for each behavior. The trajectories obtained from the simulator was discretized into cells of size $l\times b$ where $l$ is the length of the car, and $b$ is the width of each lane. With this discretization, we were able to cluster both behaviors in the demonstrations and learn their individual reward functions. The non-parametric behavior clustering algorithm was able to get a demonstration likelihood of $0.041$ while the vanilla MaxEnt IRL could only get a likelihood of the order $10^{-6}$ (the value of the objective function after convergence). The experiments took $0.13$ secs per iteration for MaxentIRL and $0.81$ secs for BCIRL (on average).  We expected to get $2$ clusters each with aggressive and evasive behaviors. However, there were five clusters each containing $21$, $19$, $5$, $4$, and $1$ demonstrations. The first two clusters correspond to evasive and aggressive behaviors while the remaining clusters did not belong to any of these behaviors. In fact, this was because these demonstrations were not consistent with the rest in any way due to the use potential field method used to generate them. This is a very important application of Nonparametric BCIRL -- though there are inconsistent demonstrations, the algorithm learns from the ones that are consistent with respect to a reward function automatically.

\section{Conclusion and Future Work}
We presented the non-parametric Behavior Clustering \ac{irl} algorithm which uses non-parametric Bayesian clustering to autonomously detect the number of underlying clusters of behavior and simultaneously cluster demonstrations generated by different behaviors while learning their reward functions. The algorithm is able to cluster different behaviors successfully as long as there exist consistent differences even though several parts of their trajectories are similar. One of the main challenges in extending this method to an arbitrary domain is choosing a feature space such that it has different feature expectations for visibly different behaviors. One possible direction is to find these features using feature construction for \ac{irl} \cite{levine2010feature}. Besides, it is noted that although discretizing continuous trajectories works well for distinguishing driving style tasks; it does not produce the underlying reward functions for the optimal control problems in the continuous domain. This points towards an extension that integrates \ac{irl} for continuous systems
such as Guided Policy Search \cite{levine2013guided} and Continuous Inverse Optimal Control \cite{levine2012continuous}.  Since this method is made computational more efficient using stochastic \ac{em} and resampling, it is applicable to urban planning problems with large data sets, \eg, transit route selection prediction \cite{wu2017datadriven}, and robot learning from demonstration from experts with different levels of performance.

\appendix 
\section{Appendix: Proof of Convergence}
In this section, we show that taking only one step of gradient ascent does not affect the convergence behavior of \ac{em}. 

First, note that each term in the original objective function given by \eqref{exact} can be written as, 

\begin{align*}
	\arg \max_w \ln p(\tau^i|w) = \arg \max_w & \underbrace{\sum_y q(y)\ln\frac{p(\tau^i,y | w)}{q(y)} }_{\text{Likelihood} = \mathcal{L}(w)}\\+ & \underbrace{\sum_y q(y) \ln\frac{q(y)}{p(y|\tau^i,w)}}_{\text{KL divergence}}
\end{align*} where, $w$ is the all the parameters we are optimizing over ($\Theta$ and $\Psi$ here) and $q(y)$ is any probability distribution $q$ over any variable $y$. We will use the fact that the KL divergence is a metric which is $0$ if and only if $q(y) = p(y | \tau^i,w)$ and is strictly greater than $0$ otherwise for the proof of convergence of one step maximization in EM. In our problem, we have $y$ as the class variable $c_j$.

\subsection{The Algorithm}
\textbf{E-step}: Given
$w^{(t)}$, the value of parameter $w$ at iteration $t$, set
$q_t(y) = p(y|\tau^i, w^{(t)})$. This makes the KL divergence at
iteration $t$ go to $0$. The log likelihood is now given by,
\begin{align*}
	\label{expectation}
	\mathcal{L}_t(w) = &\sum_j p(c_j|\tau^i, w^{(t)}) \ln p(\tau^i,c_j|w) dy \\ &- \underbrace{\sum_j p(c_j|\tau^i, w^{(t)}) \ln p(c_j|\tau^i, w^{(t)}) dy}_{\text{Independent of } w}
\end{align*}
\textbf{M-step}: Set $w^{(t+1)} = w^{(t)} + l \cdot \nabla_w\mathcal{L}_t(w)$ where, $l$ is the learning factor. This is the gradient ascent update.
Lastly , we show the algorithm converges:
\begin{align*}
	\ln p(\tau^i|w^{(t)}) &= \mathcal{L}(w) + \underbrace{KL \Big(q_t(c_j) || p(c_j | \tau^i,w^{(t)})\Big)}_{= 0\text{ by setting }q_t = p \text{ in E-step}}\\
	&= \mathcal{L}_t(w) \ \ \ \ \  \ \ \ \ \ \leftarrow \textbf{E-step}\\
	&\le \mathcal{L}_t(w^{(t+1)}) \ \        \leftarrow \textbf{M-step} \text{ (gradient ascent)}\\
	&\le \mathcal{L}_t(w^{(t+1)}) + \underbrace{KL \Big(q_t(c_j) || p(c_j | \tau^i,w^{(t+1)})\Big)}_{> 0\text{ because q }\ne p}\\
	&= \ln p(\tau|w^{(t+1)}) 
\end{align*} This shows that we still have the monotonic increase guarantee and hence convergence to at least one of the local minima. The M-step is not completely gradient ascent as shown here since we completely maximize with respect to the parameters $\Psi$. Nonetheless, as long as the likelihood is increased (for the inequality to hold) in the M-step, convergence is ensured.

\bibliographystyle{aaai}  
\bibliography{refs}  

\end{document}